\documentclass{article}

\usepackage{graphicx}
\usepackage{amsmath}
\usepackage{geometry}
\begin{document}

\title{\textbf{Acquisition of Visual Features Through Probabilistic Spike-Timing-Dependent Plasticity}}

\author{Amirhossein Tavanaei:\footnote{tavanaei@louisiana.edu} \\ \textit{\small{The Center for Advanced Computer Studies, University of Louisiana at Lafayette,}} \\ \textit{\small{Lafayette, LA 70504, USA}} \\
Timoth\'{e}e Masquelier:\footnote{timothee.masquelier@alum.mit.edu} \\ \textit{\small{INSERM, U968, Paris, F-75012, France}}\\
\textit{\small{Sorbonne Universit\'{e}s, UPMC Univ Paris 06, UMR-S 968, Institut de la Vision,}} \\ \textit{\small{Paris, F-75012, France}}\\ 
\textit{\small{CNRS, UMR-7210, Paris, F-75012, Paris, France}} \\ \textit{\small{CERCO UMR 5549, CNRS, Univ. Toulouse 3, Pavillon Baudot CHU Purpan BP 25202,}} \\ \textit{\small{31052 Toulouse Cedex, France}} \\
 Anthony S. Maida:\footnote{maida@cacs.louisiana.edu} \\ \textit{\small{The Center for Advanced Computer Studies, University of Louisiana at Lafayette,}} \\ \textit{\small{Lafayette, LA 70504, USA}}}

\date{}

\maketitle

%

\footnote{\footnotesize \textcopyright 2016 IEEE. Personal use of this material is permitted.
  Permission from IEEE must be obtained for all other uses, in any current or future
  media, including reprinting/republishing this material for advertising or promotional
  purposes, creating new collective works, for resale or redistribution to servers or
  lists, or reuse of any copyrighted component of this work in other works.}

\begin{abstract}
This paper explores modifications to a feedforward five-layer spiking convolutional network (SCN) of 
the ventral visual stream [Masquelier, T., Thorpe, S., Unsupervised learning of visual features
through spike timing dependent plasticity. PLoS Computational Biology, 3(2), 247-257]. 
The original model showed that a spike-timing-dependent plasticity (STDP) learning algorithm
embedded in an appropriately selected SCN could perform unsupervised feature discovery.
The discovered features where interpretable and could effectively be used to perform 
rapid binary decisions in a classifier.

In order to study the robustness of the previous results,
the present research examines the effects of modifying some of the components of the original model.
For improved biological realism,
we replace the original non-leaky integrate-and-fire neurons with Izhikevich-like
neurons. We also replace the original STDP rule with a novel rule that has a probabilistic interpretation.
The probabilistic STDP slightly but significantly improves the performance for both types
of model neurons. 
Use of the Izhikevich-like neuron was not found to improve performance although
performance was still comparable to the IF neuron. 
This shows that the model is robust enough to handle more biologically realistic neurons. 
We also conclude that the underlying reasons for stable performance in the model are 
preserved despite the overt changes to the explicit components of the model.


\end{abstract}

\section{Introduction}
The pattern recognition functionality found in the ventral visual pathway of
the primate brain is supported by a multi-layer, spiking biological network which is
able to learn to recognize new patterns.
An early and highly influential model of this computation is found in \cite{Serre2007a}.
They introduced a hierarchy of simple/complex cells (model neurons) that incorporated
feature extraction and location invariance pooling over larger and larger receptive field sizes.
The design has been used in many subsequent deep networks and is a member of the class of convolutional networks,
that dates back to \cite{LeCun1998a} and \cite{Fukushima1980a}.

The model in \cite{Serre2007a} did not use spiking neurons and thus could not support
spike-timing-dependent plasticity (STDP) learning, which is known to occur in many parts
of the mammalian brain.
Furthermore, under some conditions the recognition process in primates is so rapid that there is barely
enough time for even one spike to travel from the retina to the decision region in the inferior temporal
cortex \cite{Thorpe1996a}.
The first artificial spiking neural network (SNN) that included STDP and was designed to address the speed of
recognition observed in the human and primate brain is that of
Masquelier and Thorpe \cite{Masquelier2007a}.
Empirical data imposes the constraint that the majority of information used in recognition
is contained in the first neural spike for any neuron in the feedforward pipeline.
The paper \cite{Masquelier2007a} presented a model that demonstrated how this was computationally 
possible and the information provided by the first spike was sufficient to drive
an STDP-based feature discovery algorithm to support the recognition process.

The initial work was performed on the Caltech data set which presented images in a canonical 
viewpoint.
Later work in \cite{Kheradpisheh2015a} showed that this algorithm would scale to larger data sets which showed the
same object from multiple views and at multiple sizes.
The recognition performance compared favorably with other approaches, specifically, \cite{Serre2007a}
and \cite{Krizhevsky2012a}.

One limitation of the work in \cite{Masquelier2007a}  is that, despite the key observations underlying
the design of the model, it raises new questions about biological realism that beg to be answered.
Specifically, there are two limitations that the work presented here considers.
First, spike generation in \cite{Masquelier2007a} is performed by non-leaky integrate-and-fire (IF)
neurons.
This may not be a serious limitation, given the short time scale under which the model operates.
However, it would be prudent to examine the model's performance using a more realistic spike-generation
mechanism.
This is possible by replacing the IF neuron with a simplified Izhikevich model neuron \cite{Izhikevich2003a,Izhikevich2007a}.

The second question is why the STDP version used in \cite{Masquelier2007a} is effective.
Put differently, can we understand why this, or alternative STDP mechanisms, perform effective
feature discovery within this framework?
The work in \cite{Kheradpisheh2015a} shows that indeed that the STDP used in \cite{Masquelier2007a}
is an effective feature discovery mechanism.
Furthermore, a theory of how global properties of the network emerge through STDP was developed in~\cite{Masquelier2010a}.

The present paper takes preliminary steps to develop a probabilistic or Bayesian
variant of the original model in \cite{Masquelier2007a}.
In particular, we are inspired by the papers of Nessler et al. \cite{Nessler2009a,Nessler2013a}.
Those papers show that an appropriately chosen STDP rule with a particular probabilistic
interpretation, when embedded in an appropriately designed winner-take-all (WTA) network,
is able to approximate online expectation maximization.
Their paper also includes proof-of-concept experiments showing that implementations of
the model are able to perform pattern classification.

One drawback of the model in \cite{Nessler2013a} is that its trainable `excitatory' connections
use negative weights.
This is because the trained weights converge to the log of a probability value (explained later).
This gives an elegant probabilistic interpretation to the weights but since the logarithm of a
probability value is negative, the weights are necessarily negative.
The authors note that theoretically the weights can be shifted to positive values
by choosing a positive parameter, $c$, equal to the magnitude of the largest negative
weight in the implementation.
The present paper seeks a solution where the (excitatory) weights are positive.


Although the network in \cite{Masquelier2007a} has five layers, it can be viewed functionally
as consisting of
three feedforward-connected modules: feature extraction,
feature discovery, and classification.
The feature extraction and discovery components are duplicated across five spatial scales.
The feature discovery module uses a simple form of spike-time-dependent plasticity (STDP) based learning.
We compare the IF neuron with a simplified Izhikevich regular-spiking (RS) neuron which models the spike
generation dynamics of one type of neocortical pyramidal cell, the principle neuron in the neocortex.

The present research studies variants of this network when STDP is modified and when
the model neurons are modified.
Specifically, we modified the original implementation in \cite{Masquelier2007a}.
The motivation with respect to STDP was to obtain some insight as to why the learning rule 
is able to support feature discovery and to 
assess the effectiveness of our new
probabilistic STDP rule in feature discovery.
The motivation with regard to using different model neurons was to incrementally 
improve the biological realism of the model
and to examine the robustness of STDP-based feature acquisition with a different model neuron. 
Finally,
if our manipulations do not have an effect on the qualitative performance of the model,
either positive or negative, then we have established that the model has some degree
of robustness.



\section{Background}
STDP modifies synaptic weights in response to local timing differences between pre- and postsynaptic
spikes.
The existence of several variants of STDP within synapses is well established in biological networks \cite{Markram2011b}.
In most cases, however, there is little consensus about its larger role in learning for biological networks.
Idealized models of STDP have been studied extensively.

In recent literature,
there have been at least two approaches to addressing the question of why biological 
STDP might be effective in learning.
The first and most established approach seeks to obtain an idea of what kinds of temporal spike codes
STDP produces \cite{Song2001a,Guyonneau2005a}
and what kind of codes STDP can learn to recognize \cite{Masquelier2009a}.



The second approach assumes that the relevant biological networks implement a Bayesian computation.
Empirical studies have provided evidence that the some brain areas 
may perform a Bayesian analysis of sensory stimuli~\cite{Doya2007a, Kording2004a}. 
Also, it is claimed that the spike generation process is stochastic with exponential function upon the membrane potential~\cite{Rezende2011a, Jolivet2006a}. 
Probabilistic interpretations of STDP could form a theoretical link to the broader area of machine learning. 
Nessler et al. (2009, 2013) showed that a version of the STDP rule can perform Bayesian computation in a spiking winner-take-all (WTA) network~\cite{Nessler2009a, Nessler2013a}.

Based on the Nessler et al. (2013) STDP rule, Kappel et al. (2014) developed a generic cortical microcircuit to approximately implement the hidden Markov model (HMM) learning~\cite{Kappel2014a}. 
In addition, a modified variant of this probability-based STDP has been employed in a Gaussian mixture model (GMM) training of the HMM equipped by embedded SNNs in its states~\cite{Tavanaei2015b}.


The present work is inspired by the last approach and seeks to study a probabilistic rule
in the context of the network used in \cite{Masquelier2007a}.

\section{Network Architecture}  
\begin{figure}
\center
\includegraphics[width=9cm]{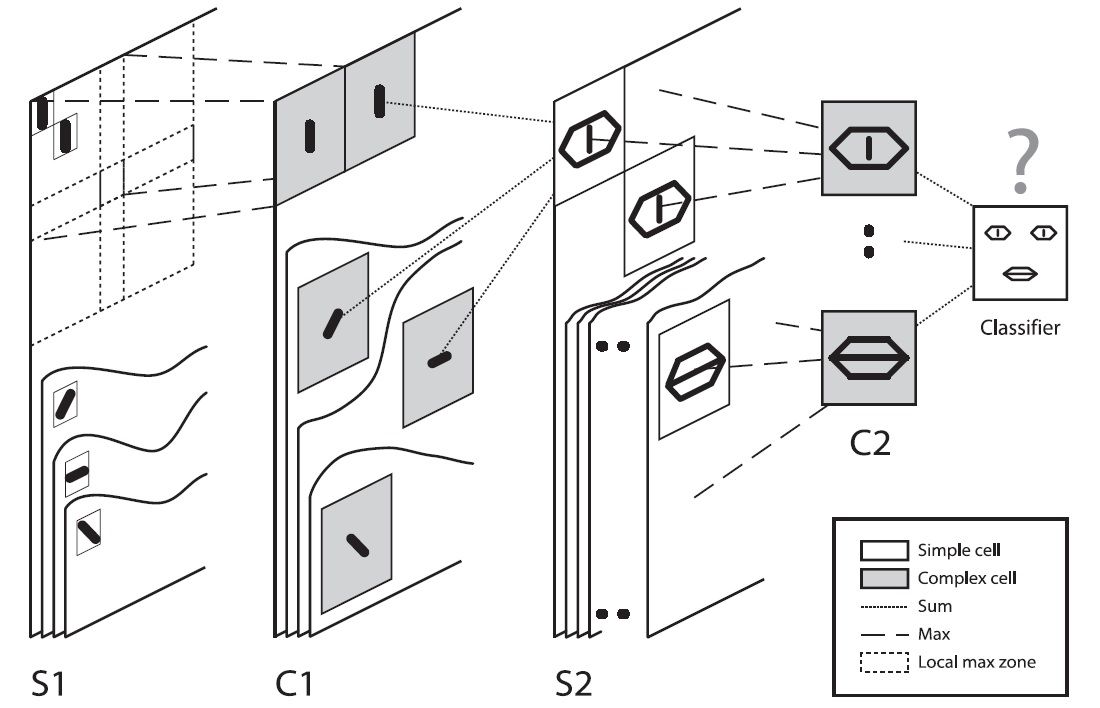}
\caption{5-layer hierarchical spiking neural network developed in \cite{Masquelier2007a}. 
There is one S1-C1-S2 pathway for each processing scale. C2 cells select the maximum response received from S2. Finally, visual features are classified by the last layer. STDP learning is applied to the synapses connecting C1 and S2 neuron layers. The synaptic weight change in the SNN proposed in~\cite{Masquelier2007a} is computed by a factor of $w(1-w)$. In the current investigation, 
we change the STDP rule and neuron models to study the effect on performance. (from \cite{Masquelier2007a})}
\label{fig:SNNArchitecture}
\end{figure}

The SNN architecture in the previous study (Masquelier and Thorpe, 2007~\cite{Masquelier2007a}) is shown in Fig.~\ref{fig:SNNArchitecture}. The network consists of four layers hierarchy of spiking neurons (S1-C1-S2-C2) in temporal domain. 
Simple cells (S) compute a linear sum (integration) while complex cells (C) compute nonlinear max pooling. 
The last layer (layer 5) performs a classification over visual features received from C2. 

Layer S1 detects edges of the image with four orientations using intensity-to-latency conversion. 
The WTA between S1 cells is implemented to find the best matching orientation. C1 maps subsample S1 maps by taking the maximum response over a square neighborhood. S2 is selective to intermediate complexity visual features, defined as a combination of oriented edges. C2 takes the maximum response of S2 over all positions and scales. The maximum operation of the complex cells propagates the first spike emitted by a given group of afferent neurons because of the time-to-first-spike framework supposed in this architecture.

STDP learning occurs in the weights projecting from layer C1 
to layer S2.
The STDP has the
effect of concentrating large synaptic weights on afferent neurons that systematically fire early, while postsynaptic spike latencies decrease. 
The synaptic weights are updated and duplicated at all positions and scales. The synaptic weights are considered between 0 and 1 to ensure excitatory synapses. A simplified STDP has been used in this structure in which the time window is supposed to cover the whole spike wave.

When a cell fires,
local inhibition 
prevents other cells in that layer from firing at the same scale and within a square neighborhood of the firing position.

The final membrane potential measured in the C2 cells (and is contrast invariant) is used for training the radial basis function (RBF) classifier. 
Images are processed by the network one by one and the neurons in each layer are allowed to fire only once. Finally, the classifier layer utilizes RBF classification to recognize the patterns according to the visual features received from C2.   

\section{The STDP rules tested}
In a later section, we will examine network classification performance using a novel STDP rule.
Here, we explain the original STDP rule used in \cite{Masquelier2007a}, along
with the new rule.
Neither rule needs to compute the exact time difference between pre- and postsynaptic spikes.

\subsection{The original STDP rule}
The original STDP rule given in \cite{Masquelier2007a} is shown below.
$w_{ki}$ denotes the weight from presynaptic neuron $i$ to postsynaptic neuron $k$.
\begin{equation}
\Delta w_{ki}=
\begin{cases}
&a^{+}\cdot w_{ki}\cdot(1-w_{ki}), \ \ t_{k}^\mathrm{f} - t_{i}^\mathrm{f} \ge 0\\
&-a^{-}\cdot w_{ki}\cdot(1-w_{ki}), \ \ \mathrm{otherwise}\\
\end{cases}
\label{MasquelierThorpeSTDP}
\end{equation}

\noindent
The first case describes LTP and the second describes LTD\@.
$t_{i}^\mathrm{f}$ and $t_{k}^\mathrm{f}$ denote firing times of units $i$ and $k$, respectively.
Weights fall in the range (0, 1).
Note that the quantity $w_{ki}\cdot (1-w_{ki})$ is the slope of the sigmoidal function.
Therefore, if $a^{+} = 1$ and $a^{-}=1$, weight changes will be largest when $w_{ki}$ is $0.5$ and weight changes 
will become arbitrarily small as $w_{ki}$ approaches the bounds 0 or 1.
In the most of simulations, the magnitudes of amplification parameters $a^{+}$ and $a^{-}$ are kept in a ratio of 4/3.

\subsection{Probabilistic STDP variant}
The proposed new rule yields weights with a probabilistic interpretation after convergence. 
Our new rule is modified from 
\cite{Nessler2013a}. 
The rule from \cite{Nessler2013a}, when embedded in an appropriate winner-take-all network with Poisson spiking
neurons can approximate expectation maximization learning.
It is given below.
\begin{equation}
\Delta w_{ki}=
\begin{cases}
&e^{-w_{ki}}-1, \ \ 0< t_{k}^\mathrm{f} - t_{i}^\mathrm{f} < \epsilon\\
&-1, \ \ \ \ \ \ \ \ \ \ \mathrm{otherwise}\\
\end{cases}
\label{nessler}
\end{equation}

\noindent
LTP occurs if the presynaptic neuron fires briefly (e.g., within $\epsilon=10$ ms) before the postsynaptic neuron.
Otherwise LTD occurs.
The rule is unusual in that the magnitude of the LTP weight adjustment is exponentially related to the current
value of the weight.
Additionally, weight values are always negative.
After convergence, the weight value is equal to the log of the probability
that the presynaptic neuron will have fired within the $\epsilon$ interval, given that the postsynaptic neuron has fired.

Our SNN uses positive weights for the excitatory neurons in the feedforward layers and therefore the 
rule~(\ref{nessler}) is not applicable. 
Therefore, we modified (\ref{nessler}) to obtain (\ref{AT_STDP}) below.

\begin{equation}
\Delta w_{ki}=
\begin{cases}
&a^+\cdot e^{-w_{ki}}, \ \ t_{k}^\mathrm{f} - t_{i}^\mathrm{f} \ge 0, \ \ \ (\mathrm{LTP})\\
&-a^-, \ \ \ \ \ \ \ \ \ \mathrm{otherwise}, \ \ \ \ (\mathrm{LTD})\\
\end{cases}
\label{AT_STDP}
\end{equation}
where $a^+$ and $a^-$ are the amplification parameters used in (\ref{MasquelierThorpeSTDP}). 
The synaptic weights are constrained to be positive during the training process. 

%

\subsection{Probabilistic perspective}
To interpret the behavior of the STDP rule probabilistically, 
we follow the approach of Nessler et al.\ (2013), p.\ 21, formula 28~\cite{Nessler2013a}. 
At equilibrium, the expected weight change will be zero. 
This can be written and then simplified 
as shown below.
The notation $z_k=1$ means that the postsynaptic neuron fired on a given stimulus presentation.
The notation $y_k=1$ means that the presynaptic neuron fired before the postsynaptic neuron on the same stimulus presentation.
\begin{equation}
\begin{array}{l}
E[\Delta w_{ki}]=0
\Leftrightarrow\\
a^+ \cdot p_\mathbf{w}^{*}(y_i = 1 | z_k = 1) e^{-w_{ki}} - a^- \cdot p_\mathbf{w}^{*}(y_i = 0 | z_k = 1) = 0
\Leftrightarrow\\
a^+ \cdot p_\mathbf{w}^{*}(y_i = 1 | z_k = 1) e^{-w_{ki}} + a^-\cdot p_\mathbf{w}^{*}(y_i = 1 | z_k = 1) = a^- 
\Leftrightarrow\\
e^{-w_{ki}}=\frac{a^- -a^- \cdot  p_\mathbf{w}^{*}(y_i = 1 | z_k = 1)}{a^+ \cdot  p_\mathbf{w}^{*}(y_i = 1 | z_k = 1)}
\Leftrightarrow\\
w_{ki}=\ln \big(\frac{a^+}{a^-} \cdot \frac{p_\mathbf{w}^{*}(y_i = 1 | z_k = 1)}{1-p_\mathbf{w}^{*}(y_i = 1 | z_k = 1)} \big) \Leftrightarrow \nonumber
\end{array}
\end{equation}
\begin{equation}
\label{eqn:equilibrium}
w_{ki}=\ln \big (\frac{a^+}{a^-}\big ) + \ln(\textit{\textbf{Odds Ratio}})
\end{equation}
where, $p^*$ denotes an equilibrium probability.

The synaptic weight, $w_{ki}$, in (\ref{eqn:equilibrium}) specifies the log odds (logit) of the probability. The probability, $p^*$, denotes the casual effect of presynaptic neuron in postsynaptic spike in which the STDP rule is triggered. 
The odds ratio is the ratio of the probability that the event of interest occurs to the probability that it does not~\cite{Bland2000a}. 
The event of interest is the presynaptic activation right before the postsynaptic neuron fires. 
Positive log odds ($p_\mathbf{w}^{*}(y_i = 1 | z_k = 1) \geq \frac{a^- }{a^- +a^+}$) specifies a probable condition in which the synaptic weight is subject to LTP. The parameter $\frac{a^-}{a^- +a^+}$ determines a stochastic threshold for postsynaptic firing. The negative log odds triggers LTD until the synaptic weight reaches its minimum efficacy. 
The minimum efficacy is when the presynaptic neuron does not have an effect on 
the postsynaptic membrane potential (disconnected synapse). 

A number of previous empirical studies have provided evidence that Bayesian analysis of sensory stimuli occurs in the brain~\cite{Doya2007a, Kording2004a, Rao2002a}. In Bayesian inference, the hidden causes are inferred using prior knowledge and the likelihood of the observations to obtain a posterior probability. In the case of two classes $M_1$ and $M_2$, the posterior probability of class $M_1$ is introduced by the sigmoid function as follows~\cite{Bishop2006a}
\begin{equation}
p(M_1|x)=\frac{1}{1+e^{-a}}
\end{equation}
where $a$ is
\begin{equation}
a=\mathrm{ln} \frac{p(x|M_1)p(M_1)}{p(x|M_2)p(M_2)}
\end{equation}
which is interpreted as log odds by
\begin{equation}
a=\mathrm{ln} \frac{p(M_1|x)}{p(M_2|x)}=\mathrm{ln} \frac{p(M_1|x)}{1-p(M_1|x)}
\label{eq:oddBishop}
\end{equation}
Equation~(\ref{eq:oddBishop}) is analogous to the synaptic weight in (\ref{eqn:equilibrium}).

\subsection{Limitations and convergence}

We consider the maximum possible weight value after a series of the LTP events during training. 
Does an excitatory synaptic weight converge to some constant or grow without bound as a function of LTP events? 
To address this, we start with initial weight, $w^0$. The next synaptic weights after performing $n$ LTP rules are $w^1, w^2, ..., w^n$. According to~(\ref{AT_STDP}),
\begin{equation}
w^{i+1}=w^i+a^+\cdot e^{-w^i}
\label{Convergence2}
\end{equation}
The maximum weight change happens when $w^i=0$. So, if we suppose $w^0=0$ (maximum change), the final synaptic weight is obtained by:
\begin{equation}
\label{Wn1}
\begin{array}{l}
w^n=w^0+a^+\cdot e^{-w^0}+...+a^+\cdot e^{-w^{n-1}}=a^+\cdot \sum_{i=0}^{n-1}e^{-w^i}
\end{array}
\end{equation}
And to find an upper bound for the synaptic weight in $n$ LTP operations, we have:
\begin{equation}
\label{upperbound1}
\begin{array}{l}
w^1=a^+, \\ 
w^2=a^+ + a^+\cdot e^{-a^+} < 2a^+,\\
w^3 < 2a^++a^+\cdot e^{-2a^+} < 3a^+, \\ 
and  ... \\ 
w^{n-1} < (n-2)a^++a^+\cdot e^{-(n-2)a^+} < (n-1)a^+ 
\end{array}
\end{equation}  
Therefore,
\begin{equation}
w^n < a^+\cdot \sum_{i=0}^{n-1}e^{-ia^+}=a^+\cdot \frac{1-e^{-na^+}}{1-e^{-a^+}}
\label{upperbound2}
\end{equation}
Finally, a synaptic weight after infinite iterations of the LTP 
adjustments (with $0<a^+<1$) will fall in range $(0, 1.582)$ where, $1.582$ is an upper bound if $a^+ = 1$.

\section{Izhikevich's Neuron Model}
Izhikevich's model for 
the RS spiking neuron concisely, but accurately, captures the spike-generation dynamics of one subtype of pyramidal neuron~\cite{Izhikevich2003a, Izhikevich2007a}. 
We used a simplified version of this so that it would be compatible with the SNN implementation
(We only use the first generated spike).
Its spike-generation dynamics is specified in Equation Set~(\ref{eqn:Izhikevich1}) using state 
variables $V$ and $U$. 

\begin{subequations}
\label{eqn:Izhikevich1}
\begin{align}
C\frac{dV}{dt}&=k(V-V_{\mathrm{rest}})(V-V_{\mathrm{th}})-U+I_{\mathrm{tot}}\\
\frac{dU}{dt}&=a[b(V-V_{\mathrm{rest}})-U]
\end{align}
\end{subequations}
$V$ denotes the membrane potential and $U$ specifies a recovery factor and keeps the membrane potential near
the resting value, $V_{\mathrm{rest}}$. 

%
%
The many different parameters:
capacity ($C$), threshold ($V_{\mathrm{th}}$), $a$, $b$, $c$, $d$, and $k$  customize the spike-generation dynamics. 
The model parameters to obtain an RS spiking neuron are: $a=0.03, b=-2, U_0=0, 
V_\mathrm{rest}=0, V_\mathrm{th}=0.49, C=100, K=0.7$; where $V_\mathrm{th}$ and $V_\mathrm{rest}$ were changed according to the model requirements.

The spike time occurs when $V$ reaches $V_\mathrm{th}$.
$I_{\mathrm{tot}}$ is set to the net synaptic input. 


\section{Experiments and Results}
We performed two experiments.
Both experiments compared four network architectures.
Specifically, we studied four types of the 5-layer SNNs as follows:
\begin{itemize}
\item{\textbf{SNN 1:} The original 5-layer convolutional network of IF spiking neurons in~\cite{Masquelier2007a} 
with the original parameters.}
\item{\textbf{SNN 2:} The SNN equipped with our probabilistic STDP learning rule.}
\item{\textbf{SNN 3:} The original SNN using Izhikevich-like neurons instead of IF neurons. 
}
\item{\textbf{SNN 4:} The SNN using Izhikevich-like neurons and equipped with the probabilistic STDP rule.}
\end{itemize}
In all of the SNNs, the learning parameter $a^+=2^{-6}$ at the beginning and is multiplied by $2$ every $400$ postsynaptic spikes, until a maximum value of $2^{-2}$. $a^-$ is adjusted such that $a^+/a^-=4/3$.

\begin{figure}[h]
\center
\includegraphics[width=8.5cm]{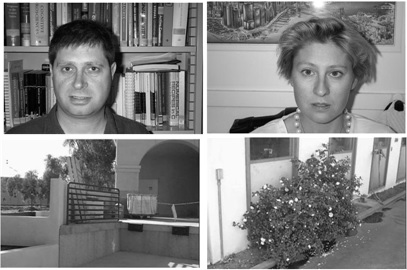}
\caption{Sample images from the Caltech dataset. 
The top row shows face examples and the bottom row shows background examples.}
\label{fig:noNoisesamples}
\end{figure}

\subsection{Experiment 1}
\subsubsection{Method}
To compare with the original work \cite{Masquelier2007a},
we evaluated the SNN using a subset of the Caltech dataset 
(http://www.vision.caltech.edu/archive.html) containing faces (positive class) 
and background images (negative class) that was used in the original model. 
We also compared motorbikes to background images.
Images were converted to gray-scale as described in \cite{Masquelier2007a}.
Examples images for face and background appear in Fig.~\ref{fig:noNoisesamples} 
(more examples, including motorbikes, are given in \cite{Masquelier2007a}). 
The dataset was equally divided into training and testing sets. 

In one simulation,
the SNNs were applied on the face/background (non-face) training samples.
In a second simulation, the bike/background data was used.

These simulations produced ten class-specific C2 features.
Each C2 cell represents a combination of edges in respect to an input image. 
A representation of the input images can be reconstructed by convolving the weight matrix with a set of kernels representing oriented bars~\cite{Masquelier2007a}. 
During the training process, initial iterations do not show a clear reconstructed preferred stimulus. 
After training, the C2 features are able to represent the face images. 
Thus, the C2 features developed selectivity to face features. 
Learning was stopped
after 1000 iterations while the weight matrices were saved after each 100 iterations to track the learning process.

The trained network was evaluated using the testing set. 
Following \cite{Masquelier2007a}, we used two measures for evaluating the model: 
1) the accuracy rate when the false positive rate equals the missed rate (equilibrium point); and,
2) A term of area under the receiver operator characteristic (ROC).

\subsubsection{Results and discussion}
\begin{figure}[h]
\center
\includegraphics[width=7cm]{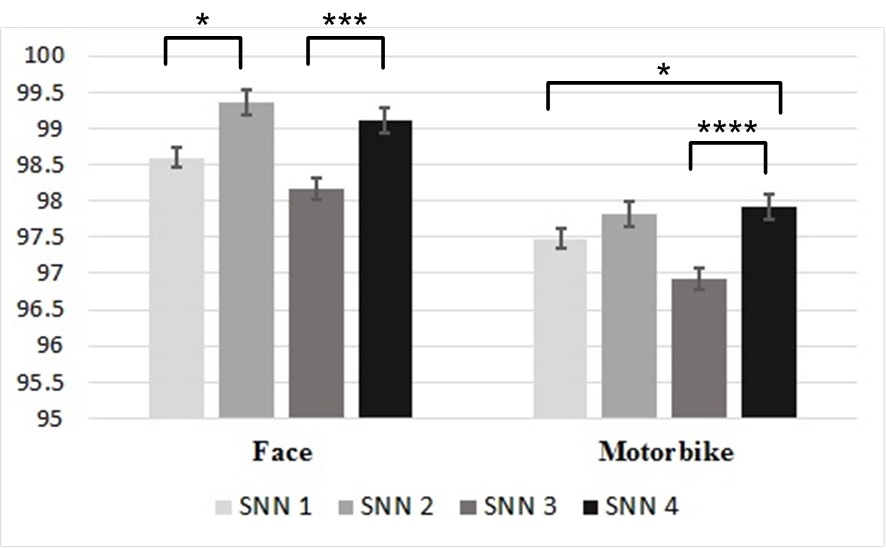}
\caption{Summary of performance for face and bike recognition in term of accuracy rate in equilibrium point (\%).}
\label{fig:barchartsAcc}
\end{figure}

\begin{figure}[h]
\center
\includegraphics[width=7cm]{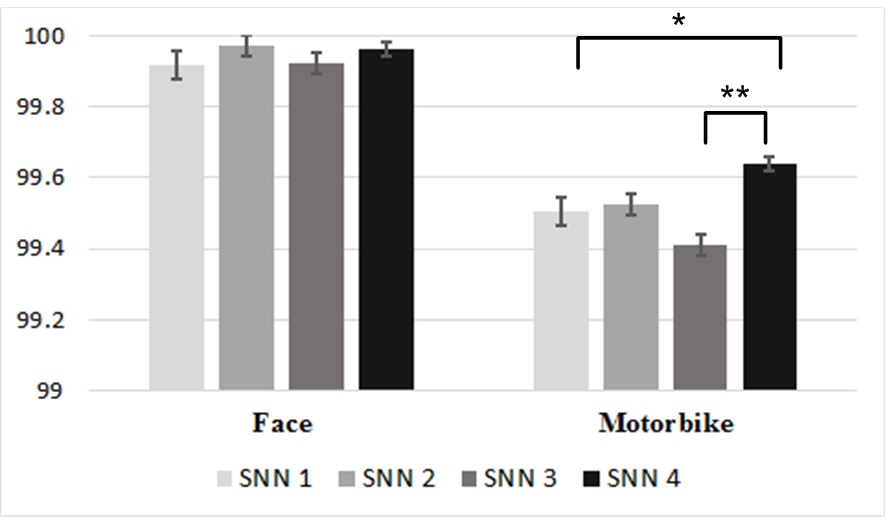}
\caption{Summary of performance for face and bike recognition in term of area under the ROC curve (\%).}
\label{fig:barchartsROC}
\end{figure}

Fig.~\ref{fig:barchartsAcc} shows the maximum accuracy rate obtained from the SNN models (\textbf{SNN 1} through \textbf{SNN 4}). 
It also includes standard errors of the mean.
These were computed by running each simulation nine times ($n=9$) with a different
set of initial values for the weights.
Different subsamples for training and testing data were not used in this first experiment
but they were used in second experiment.
Fig.~\ref{fig:barchartsAcc} shows a trend that the probabilistic STDP rule gives better accuracy 
regardless of whether the IF neuron (\textbf{SNN1} vs \textbf{SNN2}) or 
the Izhikevich-like neuron is used (\textbf{SNN3} vs \textbf{SNN4}). The overall performance for the motorbikes data was lower.
Fig.~\ref{fig:barchartsROC} illustrates the same pattern of results for the maximum ROC obtained from the SNN models for both face and motorbike datasets.

Two-tailed t-tests were performed to assess the statistical significance of the accuracy trends.
The following comparisons were performed: \textbf{SNN1} versus \textbf{SNN2}, \textbf{SNN3} versus \textbf{SNN4}, and  \textbf{SNN1} versus \textbf{SNN4} for both faces and motorbikes.
The results are shown in Figs.~\ref{fig:barchartsAcc} and \ref{fig:barchartsROC} in terms of $p<0.1$ (*), $p<0.05$ (**), $p<0.01$ (***), and $p<0.005$ (****).
The differences reached more significance for the network that used
Izhikevich-like neurons.
Thus, we claim that our probabilistic variant STDP rule significantly improves
performance when Izhikevich-like neurons are used,
either with faces or motorbikes.

\begin{figure}
\center
\includegraphics[width=8.5cm]{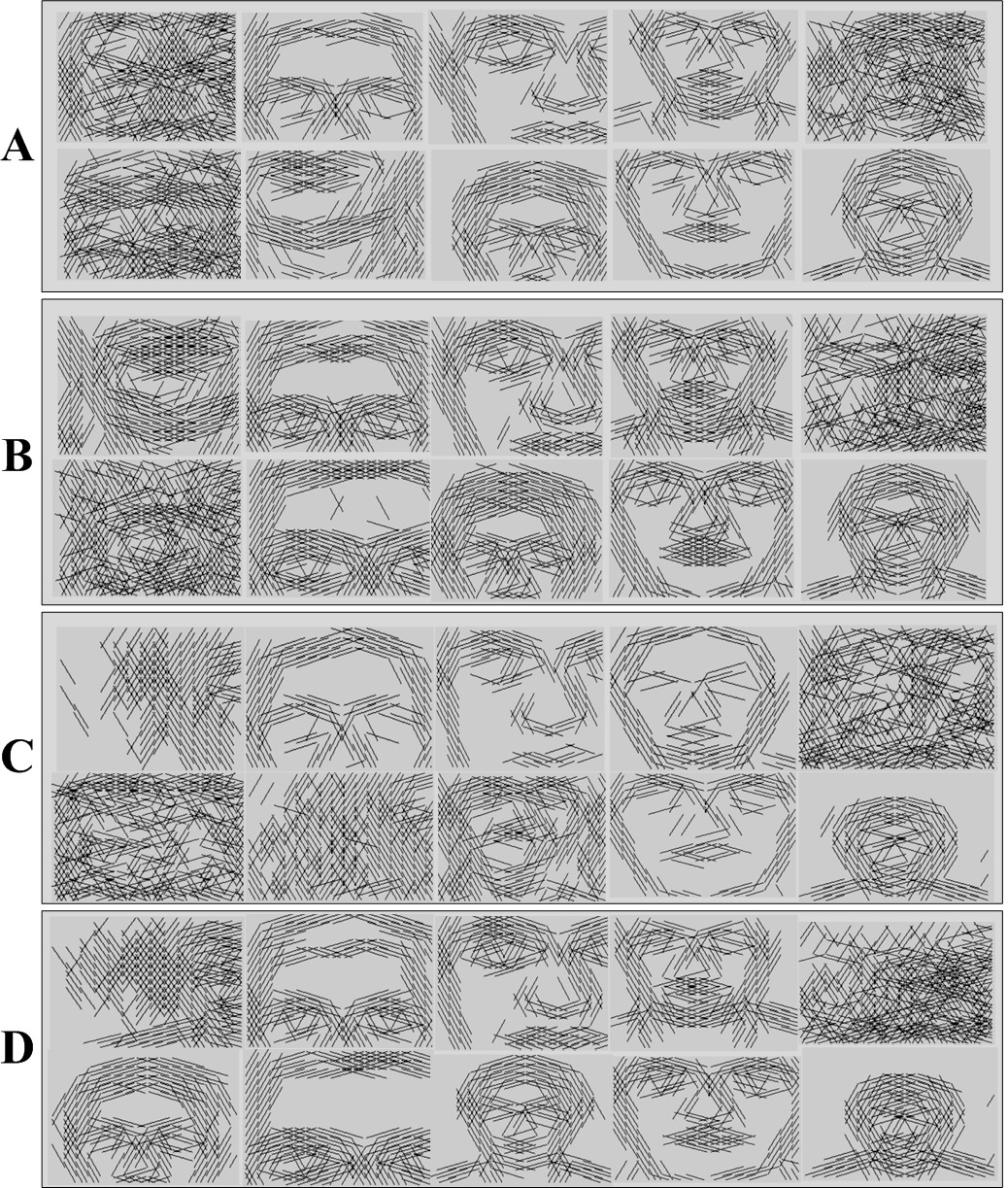}
\caption{Reconstructions for the 10 features having developed selectivity to face features. 
(A) through (D) specify the reconstructions of \textbf{SNN 1} through \textbf{SNN 4} respectively.}
\label{fig:reconstruction}
\end{figure}

\begin{figure}[h]
\center
\includegraphics[width=8.5cm]{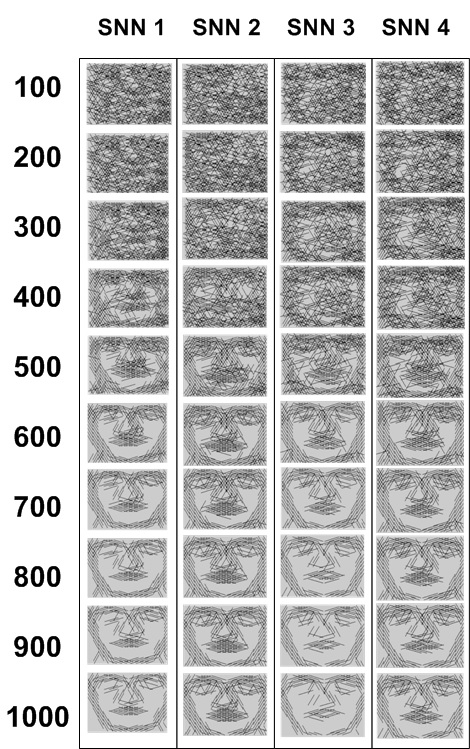}
\caption{Reconstructions of a selected face feature for \textbf{SNN 1} through \textbf{SNN4} during the training process in 100 through 1000 iterations.}
\label{fig:reconstructionIter}
\end{figure}

Fig.~\ref{fig:reconstruction}
shows reconstructions for the ten C2 cells on the face data for the four SNNs.
They indicate that face-like features were discovered by all SNN variants. 
Thus, that property of the original SNN in \cite{Masquelier2007a} is shown to be robust
under the changes introduced in this experiment.
Fig.~\ref{fig:reconstructionIter} shows the reconstructions for a selected face feature during the training process (100 to 1000 iterations) for each of the SNNs. 
Qualitatively, the results are similar for all of the SNNs.


Fig.~\ref{fig:cmp3} shows the accuracy results for all of SNNs applied to the faces data.
The most salient trend is that \textbf{SNN 3} and \textbf{SNN 4}, which both use the Izhikevich-like units,
perform better relatively early in the training, at iterations 200 and 300, but the advantage is lost later
in the training.
Also, \textbf{SNN 3} seems to perform slightly worse than the others.
Standard errors were not computed, but this was done in Experiment 2.


\subsection{Experiment 2}
The results in Experiment~1 
need further exploration
because, first, the results are
possibly compromised by ceiling effects (especially for the faces data) and, 
second, we did not perform cross-validation by using different samples
of training and testing data.

%
\subsubsection{Method}
To address the possible ceiling effect,
we used noisy test images (Gaussian noise with zero mean and standard deviation of $0.3$) for the testing phase.
The purpose was to reduce the accuracy rate so that ceiling effects were no longer an issue.
Five sets of runs ($n=5$) were performed for both faces and motorbikes.
Both testing and training images sets had 218 items.
For each simulation, 175 items were sampled from each set to be used in that simulation.
Standard errors were computed for each of 40 data points.


\subsubsection{Results and discussion}

\begin{figure}
\center
\includegraphics[width=8.5cm]{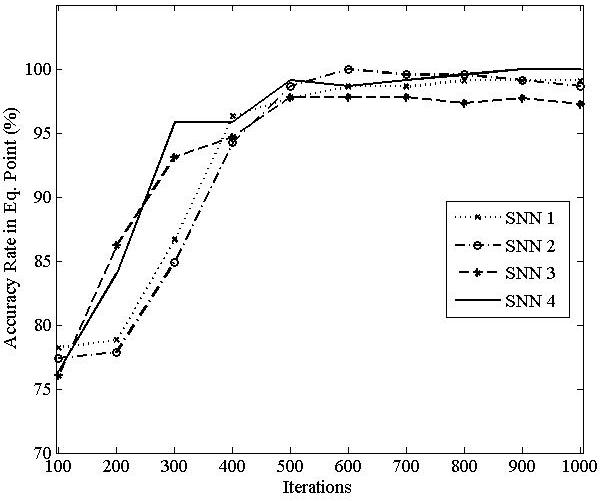}
\caption{Accuracy rate of \textbf{SNN 1} through \textbf{SNN 4} at equilibrium point in 100 through 1000 
learning iterations. (Experiment 1)}
\label{fig:cmp3}
\end{figure}

\begin{figure}
\centering
\includegraphics[width=8.5cm]{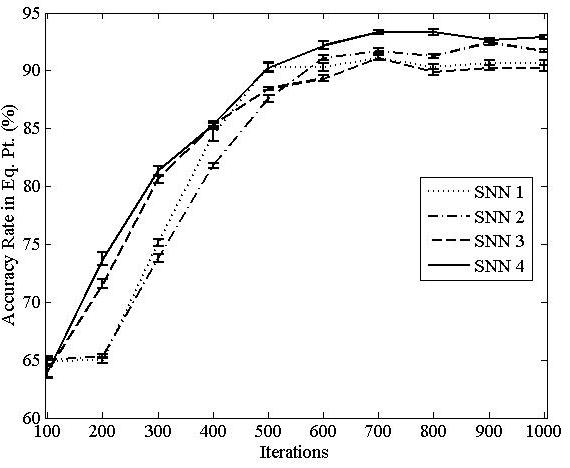}
\caption{Accuracy of face/non-face recognition for noisy images applied to \textbf{SNN 1} through \textbf{SNN 4} for 100 through 1000 learning iterations. (Experiment 2)}
\label{fig:noisyFaces_All}
\end{figure}

\begin{figure}[h]
\centering
\includegraphics[width=8.5cm]{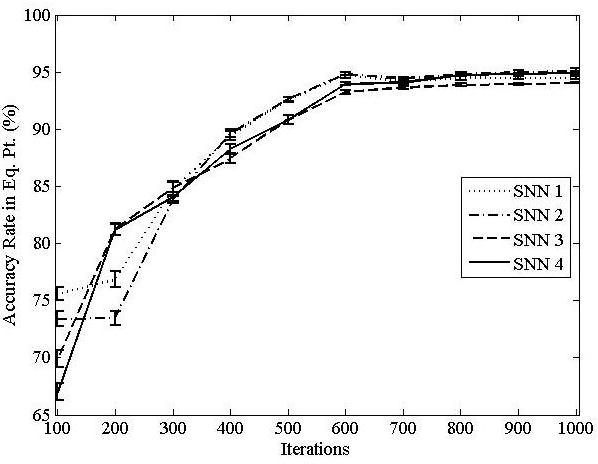}
\caption{Accuracy rates of motorbike/non-motorbike recognition for noisy images applied to \textbf{SNN 1} through \textbf{SNN 4} for 100 through 1000 learning iterations. (Experiment 2)}
\label{fig:noisyMotorbikes_All}
\end{figure}

Fig.~\ref{fig:noisyFaces_All} shows the face results for all of SNNs (\textbf{SNN 1} through \textbf{SNN 4}) including standard errors.
Fig.~\ref{fig:noisyMotorbikes_All} shows the results for motorbikes.
In both figures
the accuracy rates are below 97 percent,
so using  the noisy images eliminated the complication of a ceiling effect.
For the faces (Fig.~\ref{fig:noisyFaces_All}),
all of the trends shown in Fig.~\ref{fig:cmp3} are preserved in these results.
Most notably, the trend  where the probabilistic STDP variant improves accuracy regardless of
whether IF units or Izhikevich-like units are used still persists.
For the motorbikes, the proposed probabilistic STDP performs better; however, the accuracy performances are close across all manipulations. 
From the data taken as a whole, 
we can also conclude that the original model is robust across the manipulations.

\subsection{Control Experiments}
Experiments 1 and 2 strongly suggest that the proposed probabilistic STDP rule improves the network performance in comparison to the original one. We ran two additional control experiments to improve our confidence in this result. The first experiment controlled for weight selection and the second experiment tested several other object categories.

\begin{figure}[h]
\centering
\includegraphics[width=8.5cm]{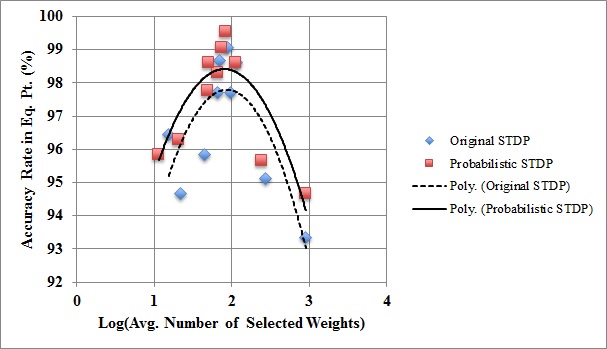}
\caption{Accuracy rates of \textbf{SNN 1} and \textbf{SNN 2} versus average number of selected weights after training. 
Data points were obtained by 10 different training runs for each SNN with $a^+/a^-=4/9, 4/8, 4/7, 4/6, 4/5, 4/4, 4/3, 4/2, 4,$ and $8$. The trend lines show better performance of the probabilistic variation of the STDP introduced in this paper.}
\label{fig:weightsvsacc}
\end{figure}

Figs.~\ref{fig:reconstruction} and~\ref{fig:reconstructionIter} show more selected  weights for probabilistic STDP (lines in face features). 
Hence, is the advantage of the probabilistic rule because it selected more weights?
To address this, ten experiments with different $a^+/a^-$ ratios were run on the face recognition task to assess the effect of the number of selected weights on accuracy. 
Fig.~\ref{fig:weightsvsacc} shows the accuracy rate versus average number of the selected weights. This figure shows that the probabilistic rule consistently outperforms the classic rule for a given number of selected weights.

\begin{figure}[h]
\centering
\includegraphics[width=10cm]{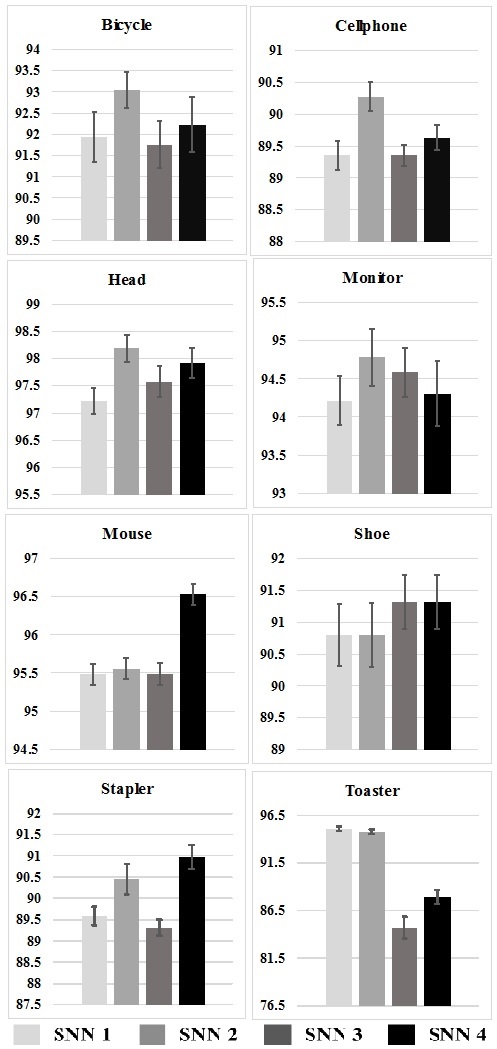}
\caption{Accuracy rates of the SNNs for recognizing eight sets of 3D-objects provided in~\cite{Savarese2007a}.}
\label{fig:3dres}
\end{figure}

Additionally, a subset of the 3D-Object images provided by Savarese et al. at CVGLab, Stanford University~\cite{Savarese2007a} was used for new experiments. 
The accuracy rates of \textbf{SNN 1} through \textbf{SNN 4} are shown in Fig.~\ref{fig:3dres}. 
This figure shows better performance of probabilistic STDP (\textbf{SNN 2} and \textbf{SNN 4}) than original one (\textbf{SNN 1} and \textbf{SNN 3} respectively) in 12 out of 16 cases.

\section{Conclusion}
This paper studied the performance of the spiking network given in
\cite{Masquelier2007a} 
when the original STDP learning rule
was replaced with a novel rule that had a probabilistic interpretation
and also 
when the model neuron was changed from an IF neuron
to an Izhikevich-like neuron. 

Our findings are the following.
First, our experiments introduced a probabilistic STDP variant which improved the accuracy rate.
Most notably, the magnitude of the weight adjustments for LTP was an exponential function
of the weight magnitude.
This is an entirely different type of rule.
Second, when we replaced the non-leaky IF neurons with Izhikevich-like neurons, accuracy performance
remained robust.
This is significant because the IF neuron is the simplest possible spike generator
whereas the Izhikevich neuron has a much more complicated spike generation mechanism.

Our main conclusion is that our probabilistic variant of STDP improved 
performance for instances of the model that used either IF neurons or Izhikevich-like neurons.
A secondary conclusion is that the original model is robust against more biologically realistic neurons (Izhikevich-like RS neurons).
Although Izhikevich-like neuron did not improve the performance, considering the nature of the manipulations, new knowledge has in fact
been obtained.
We are not aware of other work that has studied variations of this model.

A priori, one would expect that the most likely effect of any drastic change, like those
listed above, would impair network performance, but instead, the performance remained robust.
Future researchers experimenting with this model, for instance to build a deep
spiking network, can have some measure of confidence
that the performance of the individual S/C modules will remain robust when
placed in a larger context.

\end{document}